%% file: paper.tex
\documentclass[10pt,twocolumn,letterpaper]{article}

\usepackage{iccv}
\usepackage{times}
\usepackage{epsfig}
\usepackage{graphicx}
\usepackage{amsmath}
\usepackage{amssymb}

\usepackage[pagebackref=true,breaklinks=true,letterpaper=true,colorlinks,bookmarks=false]{hyperref}

\iccvfinalcopy 

\def\iccvPaperID{****} 

\input{tools}

\ificcvfinal\fi
\begin{document}

\title{Wavelet Convolutional Neural Networks for Texture Classification}

\author{Shin Fujieda\\
The University of Tokyo, Digital Frontier Inc.\\
{\tt\small sfujieda@graphics.ci.i.u-tokyo.ac.jp}
\and
Kohei Takayama\\
Digital Frontier Inc.\\
{\tt\small ktakayama@dfx.co.jp}
\and
Toshiya Hachisuka\\
The University of Tokyo\\
{\tt\small hachisuka@ci.i.u-tokyo.ac.jp}
}

\maketitle

\input{abstract}

\input{intro}
\input{related}
\input{method}
\input{results}
\input{discussion}
\input{conclusion}

{\small
\bibliographystyle{ieee}
\bibliography{references}
}

\end{document}

%% file: tools.tex
\usepackage{microtype}
\usepackage{amsfonts}

\usepackage{suffix}
\usepackage{ifthen}
\usepackage{listings}
\usepackage{paralist}
\usepackage{wrapfig}

\usepackage{algorithm,algpseudocode}
\MakeRobust{\Call}

\usepackage{color}
\definecolor{blackgreen}{rgb}{0.0,0.4,0.0}
\definecolor{gray}{rgb}{0.5,0.5,0.5}

\usepackage{bbm}

\usepackage[draft]{pdfcomment}  

\defineavatar{TODO}{color=red,open=true,opacity=0.5}
\defineavatar{SF}{author=Shin Fujieda,color=magenta,open=false,opacity=0.5}
\defineavatar{TH}{author=Toshiya Hachisuka,color=cyan,open=false,opacity=0.5}

\newcommand{\IGNORE}[1]{}

\newcommand{\Chapter}[2][]
{
	\chapter{#2}
	\ifthenelse{\equal{#1}{}}{}{\label{#1}}
}

\newcommand{\Section}[2][]
{
	\section{#2}
	\ifthenelse{\equal{#1}{}}{}{\label{#1}}
}

\newcommand{\SubSection}[2][]
{
	\subsection{#2}
	\ifthenelse{\equal{#1}{}}{}{\label{#1}}
}

\newcommand{\SubSubSection}[2][]
{
	\subsubsection{#2}
	\ifthenelse{\equal{#1}{}}{}{\label{#1}}
}

\WithSuffix\newcommand\SubSubSection*[2][]
{
	\subsubsection*{#2}
}

\newcommand{\Paragraph}[2][]
{
	\paragraph{#2}
	\ifthenelse{\equal{#1}{}}{}{\label{paragraph:#1}}
}

\newcommand{\SubParagraph}[2][]
{
	\subparagraph{#2}
	\ifthenelse{\equal{#1}{}}{}{\label{paragraph:#1}}
}

\newcommand{\Figure}[4][]
{
	\begin{figure}[htb]
		\begin{center}
			\includegraphics[width = \linewidth]{images/#2}\vspace*{-2mm}
            \ifthenelse{\equal{#4}{}}{}{\caption{#4}}
			\ifthenelse{\equal{#3}{}}{}{\label{#3}}
		\end{center}
	\end{figure}
}

\newcommand{\FigureFull}[4][]
{
	\begin{figure*}[#3]
		\begin{center}
			\includegraphics[width = \linewidth]{images/#2}
            \ifthenelse{\equal{#4}{}}{}{\caption{#4}}
			\ifthenelse{\equal{#1}{}}{}{\label{#1}}
		\end{center}
	\end{figure*}
}

\newcommand\Math[2][XXX]{%
	\begin{equation}%
		\ifthenelse{\equal{#1}{XXX}}{\nonumber}{\label{#1}}
		\begin{split}%
			#2%
		\end{split}%
	\end{equation}%
}

\newcommand\MathAlign[2][XXX]{%
	\begin{align}%
		\ifthenelse{\equal{#1}{XXX}}{\nonumber}{\label{#1}}
        #2%
	\end{align}%
}

\algnewcommand\Not{\textnormal{\textbf{not}}\xspace}

\definecolor{CommentColor}{rgb}{0,0.5,0}

\DeclareFixedFont{\mono}{T1}{fvm}{m}{n}{9pt}

\definecolor{BackgroundColor}{rgb}{1,1,1}
\definecolor{CodeColor}{rgb}{0,0,0}
\definecolor{KeyWordColor}{rgb}{0,0,0}
\definecolor{EmphasizeColor}{rgb}{0.15,0.15,0.35}

\newcommand{\SetDefaultListingParams}
{
	\lstset
	{
		captionpos=b,
		frame=none,
		language=C++,
		tabsize=2,
		lineskip=0pt,
		xleftmargin=0pt,
		xrightmargin=0pt,
        framexleftmargin=0pt,
		framexrightmargin=0pt,
        framextopmargin=0pt,
        framexbottommargin=0pt,
        columns=fullflexible,
		backgroundcolor=\color{BackgroundColor},
		basicstyle=\fontsize{7.6}{4}\sffamily\color{CodeColor},
		commentstyle=\color{CommentColor},
		keywordstyle=\bfseries\color{KeyWordColor},
		 emph={Pow, InitFrame,GetPaths,TraceRay,ProjectToEye,Render,RenderProgressive,TraceLightPaths,BuildRangeSearchStruct,IsValid,SamplePixel,Accumulate,Connect,Merge,SamplePointOnLight,SampleNextVertex,RangeSearch,TerminateRandomWalk,ContinueRandomWalk,IsEmissive,ComputeMeasurementContribution,ComputePDF,BalanceHeuristic},
        morekeywords={to,function,in},
		emphstyle={\color{EmphasizeColor}},
        mathescape=true
	}
}

\SetDefaultListingParams

\lstnewenvironment{Code}[3][]
{%
	\SetDefaultListingParams
	\ifthenelse
	{
		\equal{#1}{}
	}
	{
		\lstset
		{
			label=#2,
		}
	}
	{
		\lstset
		{
			label=#2,
			float=#1
		}
	}
	\lstset
	{
		caption={#3},
		nolol
	}
}{}

	\newenvironment{squishlist}
	{
		\begin{list}{$\bullet$}
		{
			\setlength{\itemsep}{0pt}
			\setlength{\parsep}{3pt}
			\setlength{\topsep}{3pt}
			\setlength{\partopsep}{0pt}
			\setlength{\leftmargin}{1.5em}
			\setlength{\labelwidth}{1em}
			\setlength{\labelsep}{0.5em}
		}
	}
	{
		\end{list}
	}

%% file: abstract.tex


\begin{abstract}
Texture classification is an important and challenging problem in many image processing applications.
While convolutional neural networks (CNNs) achieved significant successes for image classification, texture classification remains a difficult problem since textures usually do not contain enough information regarding the shape of object.
In image processing, texture classification has been traditionally studied well with spectral analyses which exploit repeated structures in many textures.
Since CNNs process images as-is in the spatial domain whereas spectral analyses process images in the frequency domain, these models have different characteristics in terms of performance.
We propose a novel CNN architecture, wavelet CNNs, which integrates a spectral analysis into CNNs.
Our insight is that the pooling layer and the convolution layer can be viewed as a limited form of a spectral analysis.
Based on this insight, we generalize both layers to perform a spectral analysis with wavelet transform.
Wavelet CNNs allow us to utilize spectral information which is lost in conventional CNNs but useful in texture classification.
The experiments demonstrate that our model achieves better accuracy in texture classification than existing models.
We also show that our model has significantly fewer parameters than CNNs, making our model easier to train with less memory.
\end{abstract}

%% file: intro.tex

\section{Introduction}\label{Sec:Introduction}
%
Texture is a key component used for various applications in computer graphics.
While its definition varies slightly, texture is typically a surface image of an object and it does not represent the shape of object.
For example, a photograph of an entire human face is usually not considered to be a texture, but a close-up of a human skin is.
In rendering, artists use textures to add surface details to objects without having to increase geometric complexity.
For image processing, texture is used to represent types of surfaces that are independent of shape.
Texture can be thought as a basic element that captures the appearance of surfaces of objects.

\begin{figure*}[t!]
  \centering
	\includegraphics[width=\textwidth]{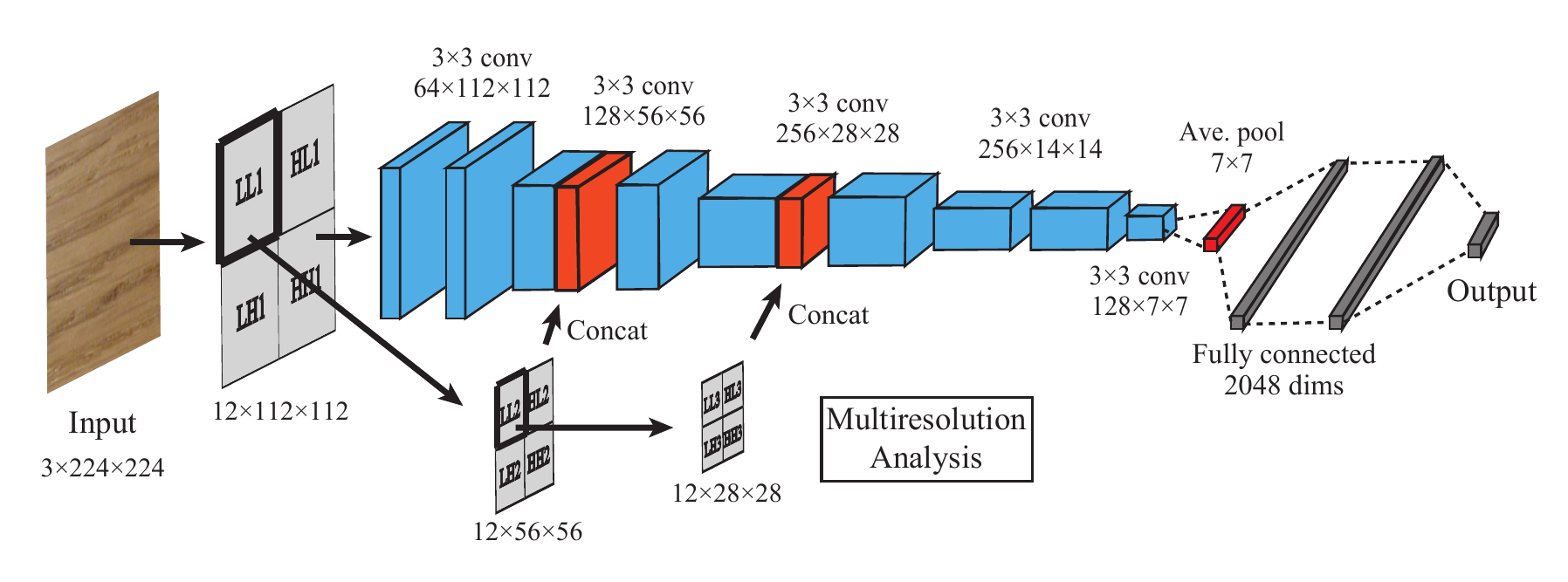}
  \vspace{-1em}
  \caption{Overview of our model with 3-level decomposition of the input image. Our model processes the input image through convolution layers with $3 \times 3$ kernels and $1 \times 1$ padding. $3 \times 3$ convolutional kernels with the stride of $2$ and $1 \times 1$ padding are used to reduce the size of feature maps. Additionally, the input image is decomposed through multiresolution analysis and the decomposed images are concatenated. The output of convolution layers is vectorized by an average pooling layer followed by three fully connected layers. The size of the output is equal to the number of classes included in the input dataset.}
  \vspace{-1em}
  \label{Fig:overview}
\end{figure*}

Accurate classification of textures is also fundamental in many important applications such as inspection and segmentation for image processing and generation of texture database for rendering.
At the same time, texture classification is a challenging problem since textures often vary a lot within the same class, due to changes in viewpoints, scales, lighting configurations, etc.
In addition, textures usually do not contain enough information regarding the shape of objects which are informative to distinguish different objects in image classification tasks.
Due to such difficulties, even the latest approaches based on convolutional neural networks achieved a limited success, when compared to other tasks such as image classification.
We propose a unification of two major classification approaches, convolutional neural networks and spectral analyses, to approach the difficulty of texture classification.
Convolutional neural networks (CNNs) process an input texture as-is and collect statistics in the spatial domain.
Spectral analysis transforms an input texture into a spectral domain and uses frequency statistics.
CNNs are usually good at capturing spatial features, while a spectral analysis is good at capturing scale invariant features.
We aim to consider both the spatial and spectral information so that it captures both types of features well under a single model.
The key idea is that the pooling layer and the convolution layer in CNNs can be thought as a limited form of a spectral analysis.
Based on this idea, we generalize both layers to perform a spectral analysis using multiresolution analysis by wavelet transform.
We thus named our model as \emph{wavelet convolutional neural networks} (wavelet CNNs).
The overview of wavelet CNNs is shown in Figure~\ref{Fig:overview}.
We demonstrate that wavelet CNNs achieve better or competitive classification accuracy while having a significantly smaller number of trainable parameters than conventional CNNs.
Our model is thus easier to train and consumes less memory than CNNs.
To summarize, our contributions are:
\begin{squishlist}
	\item Combination of CNNs and spectral analysis within one model.
	\item Generalization of pooling and convolution as a spectral analysis.
	\item Accurate and efficient texture classification using our model.
\end{squishlist}
Several numerical experiments in the results section validate that our model successfully classified failure cases of existing models.

%% file: related.tex

\section{Related Work}\label{Sec:RelatedWork}
\paragraph{Conventional Texture Descriptors}
There are a variety of approaches for extracting texture descriptors based on domain-specific knowledge.
Structural approaches~\cite{Tuceryan:1990,Goyal:1995} represent texture features based on spatial arrangements of selected pixels.
Statistical approaches~\cite{Srinicasan:2008} consider a set of statistics of pixel intensities such as mean and co-occurrence matrices~\cite{Haralick:1973,Tou:2008} as features.
These approaches can yield certain features of textures in a compact manner, but only under the assumptions in each approach.

More general approaches such as Markov random fields~\cite{Manjunath:1996}, fractal dimensions~\cite{Chaudhuri:1995}, and Wold model~\cite{Liu:1996}, described texture images as a probability model or a linear combination of a set of basis functions.
The coefficients of these models are texture features in these approaches.
While these approaches are quite general, it is still difficult choose the most suitable model for given textures.

\paragraph{Convolutional Neural Networks}
CNNs essentially replaced conventional descriptors by achieving significant performance~\cite{Krizhevsky:2012} in various tasks without relying on much of domain-specific knowledge.
For texture classification, however, directly applying CNNs is known to achieve only moderate accuracy~\cite{Hafemann:2014}.
CNNs alone thus are not very suitable for texture classification, despite its successes in other problems.

Cimpoi et al.~\cite{Cimpoi:2015} demonstrated a CNN in combination with Fisher Vectors (FV-CNN) can achieve much better accuracy in texture classification.
Their model uses a pre-trained CNN to extract texture features and this CNN part is not trained with existing texture datasets.
Inherited from conventional CNNs, this model has a large number of parameters that makes it difficult to train in general.
Our model uses a fewer parameters and achieves competitive results to FV-CNN by fusing a CNN and a spectral analysis into one model.

Andrearczyk et al.~\cite{Andrearczyk:2016} proposed texture CNN (T-CNN) which is a CNN specialized for texture classification.
T-CNN uses a novel energy layer in which each feature map is simply pooled by calculating the average of its activated output.
This results in a single value for each feature map, similar to an energy response to a filter bank.
This approach does not improve classification accuracy, but to its simple architecture reduces the number of parameters.
While our model is inspired by the problem of texture classification, it is not specialized for texture classification.
As discussed in Section~\ref{Sec:Discussion}, we confirmed that our model achieves competitive performance to AlexNet~\cite{Krizhevsky:2012} for image classification.

\paragraph{Spectral Approaches}
Spectral approaches transform textures into the frequency domain using a set of spatial filters.
The statistics of the spectral information at different scales and orientations define texture features.
This approach has been well studied in image processing and achieved practical results~\cite{Unser:1995,Arivazhagan:2007,Kanchana:2013}.

Feature extraction in the frequency domain has two advantages.
First, a spatial filter can be easily made selective by enhancing only certain frequencies while suppressing others.
This explicit selection of certain frequencies is difficult to control in CNNs.
Additionally, the periodical structure of a texture can be represented as a certain spatial frequency in the spectral domain.

Figure~\ref{Fig:overview} shows that the structure of our model is similar to that of CNNs using skip-connections~\cite{Long:2015, Ronneberger:2015}.
While deep neural networks including CNNs with skip-connections are known to be universal approximators~\cite{hornik1991approximation}, it is not clear whether CNNs can learn to perform spectral analyses in practice with available datasets.
We thus propose to directly integrate spectral approaches into CNNs, particularly based on multiresolution analysis using wavelet transform~\cite{Unser:1995}.
Our experiments support this observation that a CNN with more parameters cannot be trained to become equivalent to our model with available datasets in practice.
The key difference is that certain layers of wavelet CNNs have no trainable parameters.
Instead, those layers perform multiresolution analysis using fixed parameters defined by wavelet transform.
%

%% file: method.tex

\section{Wavelet Convolutional Neural Networks}\label{Sec:Method}

\paragraph{Overview}
We propose to formulate convolution and pooling in CNNs as filtering and downsampling.
This formulation allows us to connect convolution and pooling with multiresolution analysis.
In the following explanations, we use a single-channel 1D data for the sake of brevity.
Applications to 2D images with multiple channels are trivially possible as was done by CNNs.

\subsection{Convolutional Neural Networks}
\begin{figure}[t]
  \centering
  \includegraphics[width=\columnwidth]{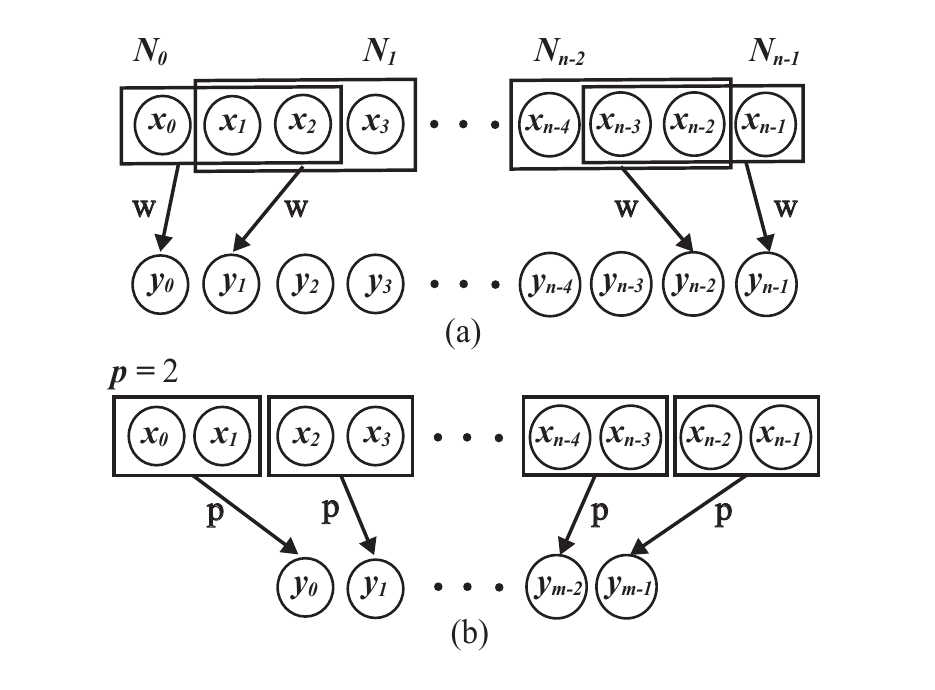}\vspace{-0.2cm}
	\vspace{-2mm}
  \caption{Concepts of convolution and average pooling layers. (a) Convolution layers compute a weighted sum of neighbors in each local receptive field. (b) Pooling layers take an average and perform downsampling.}
	\vspace{-2mm}
	\label{Fig:GenPooling}
\end{figure}

A convolutional neural network~\cite{Krizhevsky:2012} is a variant of the neural network which uses a sparsely connected deep network.
In the regular neural network model, every input is connected to every unit in the next layer.
In addition to the use of an activation function and a fully connected layer, CNNs introduce convolution/pooling layers that connect only to local neighborhoods (called a local receptive field) around each input.
Figure~\ref{Fig:GenPooling} illustrates the configuration we explain in the following.

\paragraph{Convolution Layers:}
Given an input vector with $n$ components $\mathbf{x} = (x_0, x_1, \ldots, x_{n-1}) \in \mathbb{R}^n$, a convolution layer outputs a vector of the same number of components $\mathbf{y} = (y_0, y_1, \ldots, y_{n-1}) \in \mathbb{R}^n$:
\Math[Eqn:Convolution]{
	y_{i} = \sum_{j \in N_i} w_{j} x_{j},
}
where $N_i$ is a set of indices in the local receptive field at $x_i$ and $w_{j}$ is a weight.
Following the notational convention in CNNs, we consider that $w_{j}$ includes the bias by having a constant input of $1$.
The equation thus says that each output $y_{i}$ is a weighted sum of neighbors $\sum_{j \in N_i} w_{j} x_{j}$ plus constant.

Each layer defines the weights $w_{j}$ as constants over $i$.
By sharing parameters, CNNs reduce the number of parameters and achieve translation invariance in the image space.
The definition of $y_{i}$ in Equation~\ref{Eqn:Convolution} is equivalent to convolution of $x_i$ via a filtering kernel $w_{j}$, thus this layer is called a convolution layer.
We can thus rewrite $y_{i}$ in Equation~\ref{Eqn:Convolution} using the convolution operator $\ast$ as
\Math[Eqn:Convolution2]{
	\mathbf{y} = \mathbf{x} \ast \mathbf{w},
}
where $\mathbf{w} = (w_0, w_1, \ldots, w_{m-1}) \in \mathbb{R}^m$.
Convolution layers in CNNs typically use different sets of weights for the same input and output the results as a concatenated vector.
This common practice just applies Equation~\ref{Eqn:Convolution} repeatedly for each set of weights.

\paragraph{Pooling Layers:}
Pooling layers are typically used immediately after convolution layers to simplify the information.
While max pooling is used in many applications of CNNs, Gatys et al.~\cite{Gatys:2015} showed that average pooling is more suitable for extracting texture features.
We thus focus on average pooling, which in fact allows us to see the connection with multiresolution analysis.
Given an input $\mathbf{x} \in \mathbb{R}^n$, average pooling outputs a vector of a fewer components $\mathbf{y} \in \mathbb{R}^m$ as
\Math[Eqn:Pooling]{\vspace{-1EM}
	y_j =  \frac{1}{p} \sum_{k = 0}^{p-1} x_{p j + k},
}
where $p$ defines the support of pooling and $m = \frac{n}{p}$.
For example, $p = 2$ means that we reduce the number of outputs to a half of the inputs by taking pair-wise averages.
Using the standard downsampling operator $\downarrow$, we can rewrite Equation~\ref{Eqn:Pooling} as
\Math[Eqn:Pooling2]{
	\mathbf{y} = (\mathbf{x} \ast \mathbf{p} ) \downarrow p,
}
where $\mathbf{p} = (1/p, \ldots, 1/p) \in \mathbb{R}^p$ represents the averaging filter.
Average pooling mathematically involves convolution via $\mathbf{p}$ followed by downsampling with the stride of $p$.

\subsection{Generalized Convolution and Pooling}
Equation~\ref{Eqn:Convolution2} and Equation~\ref{Eqn:Pooling2} can be combined into a generalized form of convolution and downsampling as
\Math[Eqn:GenPooling]{
	\mathbf{y} = (\mathbf{x} \ast \mathbf{k} ) \downarrow p.
}
The generalized weight $\mathbf{k}$ is defined as
\begin{squishlist}
	\item $\mathbf{k} = \mathbf{w}$ with $p=1$ (convolution in Equation~\ref{Eqn:Convolution2})
	\item $\mathbf{k} = \mathbf{p}$ with $p>1$ (pooling in Equation~\ref{Eqn:Pooling2})
	\item $\mathbf{k} = \mathbf{w} \ast \mathbf{p}$ with $p>1$ (convolution followed by pooling).
\end{squishlist}

Our insight is that Equation~\ref{Eqn:GenPooling} is equivalent to a part of a single level application of multiresolution analysis.
Suppose that we use a pair of low-pass $\mathbf{k}_{l}$ and high-pass $\mathbf{k}_{h}$ filters to decompose data into the low-frequency part $\mathbf{x}_{\mathrm{low}}$ and the high frequency part $\mathbf{x}_{\mathrm{high}}$ with $p = 2$:
\Math[Eqn:Filtering]{
	&\mathbf{x}_{\mathrm{low}} = (\mathbf{x} \ast \mathbf{k}_{l} ) \downarrow 2 \\
	&\mathbf{x}_{\mathrm{high}} = (\mathbf{x} \ast \mathbf{k}_{h} ) \downarrow 2.
}
Our key insight is that multiresolution analysis~\cite{crowley1981representation} further decomposes the low frequency part $\mathbf{x}_{\mathrm{low}}$ into its low frequency part and high frequency part by repeatedly applying the same form of Equation~\ref{Eqn:Filtering}.
By defining $\mathbf{x}_{\mathrm{low, 0}} = \mathbf{x}$, multiresolution analysis can be written as
\Math[Eqn:MRA]{
	&\mathbf{x}_{\mathrm{low},l+1} = (\mathbf{x}_{\mathrm{low},l} \ast \mathbf{k}_{l} ) \downarrow 2 \\
	&\mathbf{x}_{\mathrm{high},l+1} = (\mathbf{x}_{\mathrm{low},l} \ast \mathbf{k}_{h} ) \downarrow 2.
}
The number of applications $l$ is called a level in multiresolution analysis.
Multiresolution analysis is thus equal to repeated applications of generalized convolution and pooling layers on low-frequency parts with a specific pair of convolution kernels.

\begin{figure}[t]
	\vspace{-0.4cm}
  \centering
  \includegraphics[width=\columnwidth]{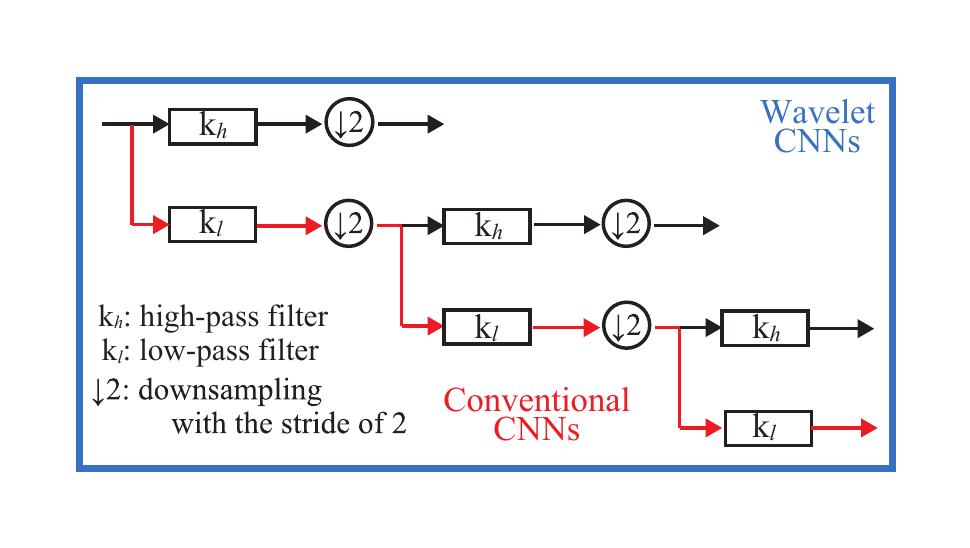}\vspace{-0.4cm}
	\vspace{-3mm}
  \caption{Relationship between conventional CNNs and wavelet CNNs in terms of multiresolution analysis. Conventional CNNs apply convolution and pooling repeatedly to the input, which is essentially equivalent to use only the low frequency components of multiresolution analysis. Our wavelet CNNs instead consider all the components including high frequency components.\label{Fig:relation}}
  \vspace{-2mm}
\end{figure}

Figure~\ref{Fig:relation} illustrates how CNNs and our wavelet CNNs differ under this formulation.
Conventional CNNs can be seen as using only the low frequency parts and discard all the high frequency parts.
Our model instead uses both the low frequency parts \emph{and} the high frequency parts within CNNs, so that we do not lose any information of the input $\mathbf{x}$ by the definition of multiresolution analysis.
While this idea might look simple after the fact, our model is powerful enough to outperform the existing more complex models as we will show in the results.

Note that we cannot use an arbitrary pair of filters ($\mathbf{k}_{l}$ and $\mathbf{k}_{h}$) to perform multiresolution analysis.
To avoid any loss of frequency information of the input $\mathbf{x}$, a pair should satisfy the quadrature mirror filter relationship.
For wavelet transform, $\mathbf{k}_{h}$ is known as the wavelet function and $\mathbf{k}_{l}$ is known as the scaling function.
We used Haar wavelets~\cite{haar1910theorie} for our experiments, but our model is not restricted to Haar.
This constraint also suggests why it is difficult to train conventional CNNs to perform the same computation as wavelet CNNs do: weights in CNNs are ignorant of this important constraint to satisfy and just try to learn it from datasets.

Rippel et al.~\cite{Rippel:2015} proposed a related approach of replacing convolution and pooling by discrete Fourier transform and truncation of the coefficients.
This approach, called \emph{spectral pooling}, is equivalent to using only the low frequency part in our model, thus it is not essentially different from conventional CNNs.
Our model is also different from merely applying multiresolution analysis on input data and using CNNs afterward, since multiresolution analysis is built inside the network and the input first go through CNN layers both before and after multiresolution analysis.

\subsection{Implementation}
\paragraph{Network Structure}
Figure~\ref{Fig:overview} illustrates our network structure.
We designed our network structure after a VGG-19 network~\cite{Simonyan:2014} since it has been successfully used for extracting features of textures for different purposes~\cite{Gatys:2015,Aittala:2016}.
We use $3 \times 3$ convolutional kernels exclusively and $1 \times 1$ padding to ensure the output is the same size as the input.

Instead of using the pooling layers to reduce the size of the feature maps, we exploit convolution layers with the increased stride.
If $1 \times 1$ padding is added to the layer with a stride of two, the output becomes half the size of the input layer.
This approach can be used to replace max pooling without loss in accuracy~\cite{Springenberg:2015}.
In addition, since both the VGG-like architecture and image decomposition in multiresolution analysis have the same characteristic that the size of images is reduced to a half successively, we combine each level of decomposed images with feature maps of the specific layer that are the same size as those images.

For texture classification, inserting an energy layer directly before fully connected layers improves the performance of a network while keeping the number of parameters small~\cite{Andrearczyk:2016}.
We used this approach and a complete network of wavelet CNNs we tested consists of nine convolution layers, the same number of wavelet layers as decomposition levels of multiresolution analysis and an energy layer followed by three fully connected layers.
We implemented this network using Caffe~\cite{Jia:2014}.
The codes for our model will be available on our website.

\paragraph{Learning}
Wavelet CNNs exploit an energy layer with the same size as the input of the layer, so the size of input images is required to be the fixed size.
We thus train our proposed model exclusively with images of the size $224 \times 224$.
These images are achieved by first scaling the training images to $256 \times 256$ pixels and then conducting random crops to $224 \times 224$ pixels and flipping.
This random variation helps the model to prevent overfitting.
For further robustness, we use batch normalization~\cite{Ioffe:2015} throughout our network before activation layers during training.
For the optimizer, we exploit the Adam optimizer~\cite{Kingma:2014} instead of SGD.
We use the Rectified Linear Unit (ReLU)~\cite{Xavier:2011} as the activation function in all the experiments.
%

%% file: results.tex
\section{Results}\label{Sec:Results}
\paragraph{Datasets}
For our experiments, we used two publicly available texture datasets: \emph{kth-tips2-b}~\cite{Hayman:2004} and \emph{DTD}~\cite{Cimpoi:2014}.
The \emph{kth-tips2-b} dataset contains 11 classes of 432 texture images.
Each class consists of four samples and each sample has 108 images.
Each sample is used for training once while the remaining three samples are used for testing.
The results for kth-tips2-b are shown as the mean and the standard deviation over the four splits.
The \emph{DTD} dataset contains 47 classes of 120 images "in the wild" which means that images are collected in uncontrolled conditions.
This dataset includes 10 available annotated splits with 40 training images, 40 validation images, and 40 testing images for each class.
The results for DTD are averaged over the 10 splits.
We processed the images in each dataset by global contrast normalization.
We calculated the accuracy as percentage of images that are correctly labeled which is a common metric in texture classification.

\begin{figure}[t]
  \centering
  \includegraphics[width=\columnwidth]{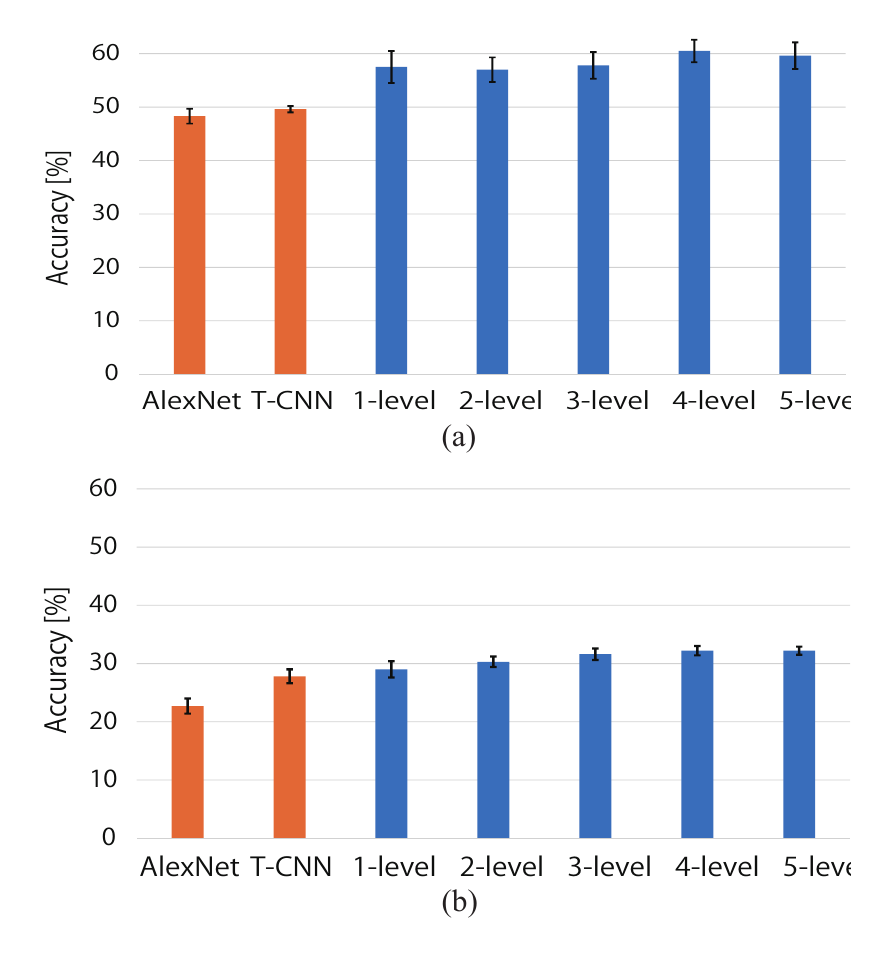}
  \vspace{-3mm}
  \caption{Classification results of (a) kth-tips2-b and (b) DTD for networks trained from scratch. We compared our models (blue) with AlexNet and T-CNN.}
  \vspace{-2mm}
  \label{Fig:accuracy_fromscratch}
\end{figure}

\begin{table*}[t]
    \centering
    \begin{tabular}{| p{1.8cm} || p{1.5cm} | p{1.5cm} | p{1.5cm} | p{1.5cm} | p{1.5cm} | p{1.5cm} | p{1.5cm} |} \hline
      & AlexNet & T-CNN & 1-level & 2-level & 3-level & 4-level & 5-level \\ \hline
      kth-tips2-b & $48.3_{\pm 1.4}$ & $49.6_{\pm 0.6}$ & $57.5_{\pm 3.0}$ & $57.0_{\pm 2.3}$ & $57.8_{\pm 2.5}$ & ${\bf 60.5_{\pm 2.1}}$ & $59.6_{\pm 2.5}$ \\ \hline
      DTD & $22.7_{\pm 1.3}$ & $27.8_{\pm 1.2}$ & $29.0_{\pm 1.4}$ & $30.3_{\pm 0.9}$ & $31.6_{\pm 1.0}$ & ${\bf 32.2_{\pm 0.8}}$ & ${\bf 32.2_{\pm 0.7}}$ \\ \hline
    \end{tabular}
    \vspace{2mm}
    \caption{Classification results for networks trained from scratch indicated as accuracy (\%).}
    \label{Table:accuracy_fromscratch}
\end{table*}

\paragraph{Training from scratch}
We compared our model with AlexNet and T-CNN using texture datasets to train each model from scratch.
For initialization of the parameters, we used a robust method that specifically accounts for ReLU~\cite{He:2015}.
While the structure of our models is designed after VGG networks, we found that VGG networks tend to perform poorly due to over-fitting with a large number of trainable parameters, if trained from scratch.
We thus used AlexNet as an example of conventional CNNs instead of VGG networks for this experiment.
Figure~\ref{Fig:accuracy_fromscratch} and Table~\ref{Table:accuracy_fromscratch} show the results of training our model with different levels of multiresolution analysis.
For both datasets, our models perform better than AlexNet and T-CNN.

Comparing between different levels within wavelet CNNs, we found that the network with 4-level decomposition performed the best.
In our experiments, the model with 5-level decomposition achieved almost the same accuracy as 4-level, but with more trainable parameters.
Similar to the number of layers in CNNs, the decomposition level of wavelet CNNs is another hyper-parameter which can be tuned for different problems.

\begin{figure}[t]
  \vspace{-1mm}
  \centering
  \includegraphics[width=\columnwidth]{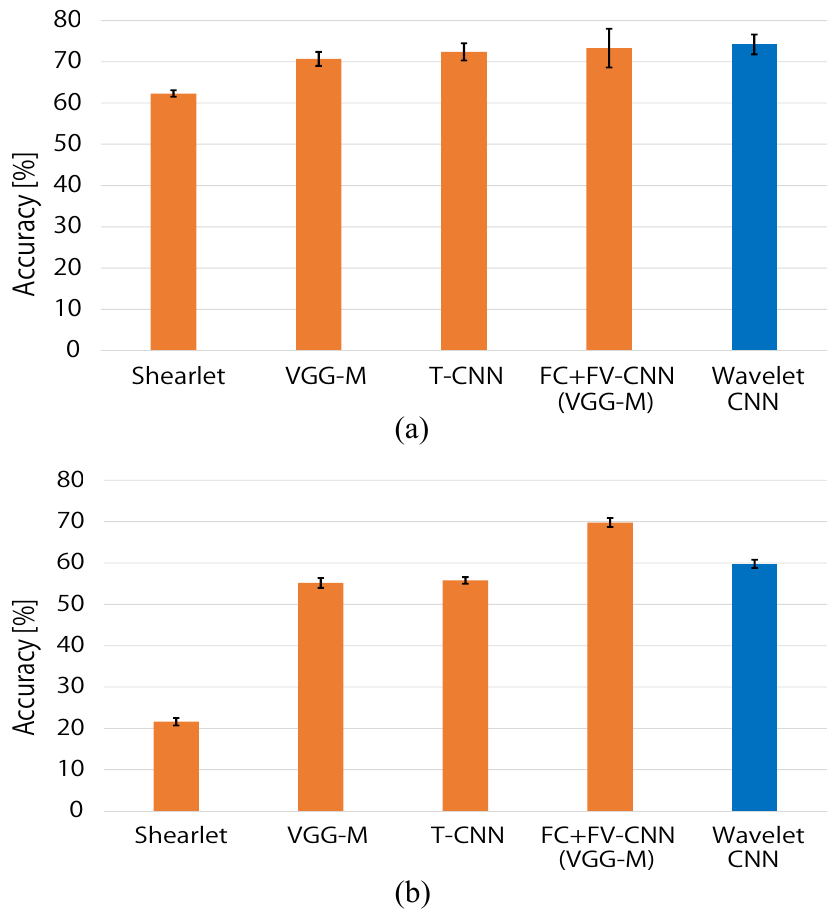}\vspace{-0.1cm}
  \vspace{-3mm}
  \caption{Classification results of (a) kth-tips2-b and (b) DTD for networks pre-trained with ImageNet 2012 dataset. We compared our model (wavelet CNN with 4-level decomposition) with shearlet transform, VGG-M, T-CNN, and FC+FV-CNN.}
  \vspace{-2mm}
  \label{Fig:accuracy_finetuned}
\end{figure}

\paragraph{Training with fine-tuning}
Figure~\ref{Fig:accuracy_finetuned} and Table~\ref{Table:accuracy_finetuned} show the classification rates using the networks pre-trained with the ImageNet 2012 dataset~\cite{Russakovsky:2015}.
Since the model with 4-level decomposition achieved the best accuracy in the previous experiment, we used this network in this experiment as well.
We compared our model with a spectral approach using shearlet transform~\cite{Krishnan:2016}, a VGG network~\cite{Chatfield:2014}, T-CNN~\cite{Andrearczyk:2016}, and FV-CNN~\cite{Cimpoi:2015} with a fully connected layer (FC).

Our model achieved the best performance for the \emph{kth-tips2-b} dataset, while it is outperformed only by FV-CNN for the \emph{DTD} dataset.
We analyzed the results and found that some classes of the \emph{DTD} dataset contain non-texture images that clearly show the shape of an object.
Since FV-CNN has a significantly larger number of trainable parameters than our models, we speculate that FV-CNN is just trained to account of for non-texture images as special cases.
We put FC-CNN as it is the state-of-the-art, a comparison with FV-CNN only on accuracy is not necessarily very fair because of this sheer difference in model complexity.

\begin{figure}[t]
  \vspace{-3mm}
  \centering
  \includegraphics[width=\columnwidth]{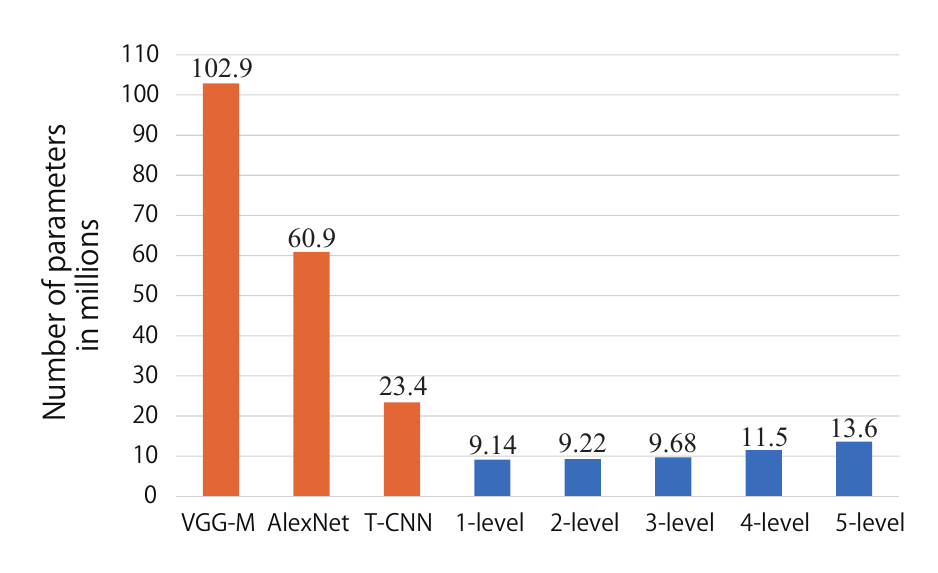}\vspace{-0.4cm}
  \vspace{-3mm}
  \caption{The number of trainable parameters in millions. Our model, even with 5-level of multiresolution analysis, has a fewer parameters than any other competing models we tested.}
  \vspace{-2mm}
  \label{Fig:parameters}
\end{figure}

\paragraph{Number of parameters}
To assess the complexity of each model, we compared the number of trainable parameters such as weights and biases for classification to 1000 classes (Figure~\ref{Fig:parameters}).
Conventional CNNs such as VGG-M (which is used also in FV-CNN) and AlexNet have a large number of parameters while their depth is a little shallower than our proposed model.
Even compared to T-CNN, which aims at reducing the model complexity, the number of parameters in our model with 4-level decomposition is less than the half.
We also remind that our model achieved higher accuracy than T-CNN does.

This result confirms that our model achieves better results with a significantly reduced number of parameters than existing models.
The memory consumption of each Caffemodel is: 392~MB (VGG-M), 232~MB (AlexNet), 89.1~MB (T-CNN), and 43.8~MB (Ours).
Our network is thus suitable to run on a system with a limited amount of memory.
The small number of parameters also generally suppress over-fitting of the model for small datasets.

\begin{table}[t]
    \centering
    \scalebox{0.8}{
      \begin{tabular}{| l || l | l | l | l | l |} \hline
        & \raisebox{0.2cm}{Shearlet} & \raisebox{0.2cm}{VGG-M} & \raisebox{0.2cm}{T-CNN }& \raisebox{-0.1cm}{\shortstack{FC+ \\ FV-CNN \\ (VGG-M)}} & \raisebox{0.1cm}{\shortstack{Wavelet \\ CNN}} \\ \hline
        kth-tips2-b & $62.3_{\pm 0.8}$ & $70.7_{\pm 1.7}$ & $72.4_{\pm 2.1}$ & $73.9_{\pm 4.9}$ & ${\bf 74.2_{\pm 1.2}}$ \\ \hline
        DTD & $21.6_{\pm 0.9}$ & $55.2_{\pm 1.2}$ & $55.8_{\pm 0.8}$ & ${\bf 69.8_{\pm 1.1}}$ & $59.8_{\pm 0.9}$ \\ \hline
      \end{tabular}
    }
    \caption{Classification results for networks pre-trained with ImageNet indicated as accuracy (\%).}
    \label{Table:accuracy_finetuned}
\end{table}

\begin{figure*}[p]
  \vspace{-5mm}
  \centering
  \includegraphics[width=\textwidth]{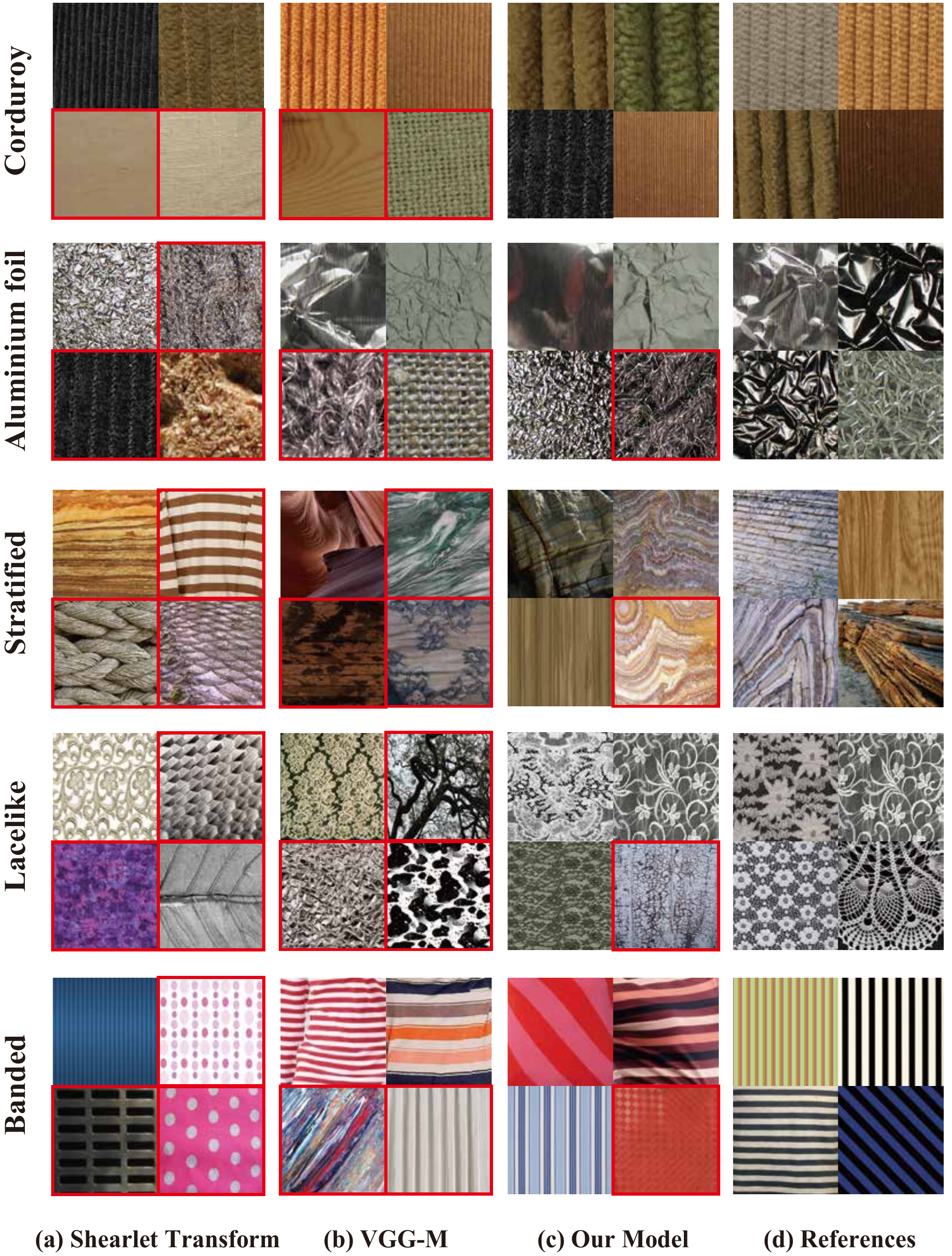}
  \vspace{-3mm}
  \caption{Some results classified by (a) shearlet transform, (b) VGG-M, (c) our model and (d) references. The images on top two rows are extracted from \emph{kth-tips2-b} and the rests are extracted from \emph{DTD}. The images in red squares are wrongly classified images.}
  \label{Fig:results}
\end{figure*}

\paragraph{Visual comparisons of classified images}
Figure~\ref{Fig:results} shows some extracted images for several classes in our experiments.
The images in top two rows of Figure~\ref{Fig:results} are from \emph{kth-tips2-b} dataset, while the images in the bottom three rows of Figure~\ref{Fig:results} are from \emph{DTD} dataset.
A red square indicates that the texture is inappropriately classified to the class.
We can visually see that a spectral approach is insensitive to the scale variation and extract detailed features, whereas a spatial approach is insensitive to distortion.
For example, in \emph{Aluminium foil}, the image of wrinkled aluminium foil is properly classified with a shearlet transform, but not with VGG-M.
In \emph{Banded}, VGG-M classifies distorted lines into the correct class.
Since our model is the combination of both approaches, it can assign texture images to the correct label in every variation above.

%% file: discussion.tex

\section{Discussion}\label{Sec:Discussion}
\paragraph{Application to image classification}
Since we do not assume anything regarding the input, our model is not necessarily restricted to texture classification.
To confirm the generality, we trained a wavelet CNN with four-level decomposition and AlexNet with the ImageNet 2012 dataset from scratch to perform image classification.
Our model obtained the accuracy of 59.8\% whereas AlexNet resulted in 57.1\%.
We should remind that the number of parameters of our model is about five times smaller than that of AlexNet (Figure~\ref{Fig:parameters}).
Our model is thus suitable also for image classification with smaller memory footprint.
Other applications such as image recognition and object detection with our model should be similarly possible.

\paragraph{$\mbox{\boldmath $L_p$}$ pooling}
An interesting generalization of max and average pooling is $L_p$ pooling~\cite{Boureau:2010,Sermanet:2012}.
The idea of $L_p$ pooling is that max pooling can be thought as computing $L_{\infty}$ norm, while average pooling can be considered as computing $L_{1}$ norm.
In this case, Equation~\ref{Eqn:Pooling2} cannot be written as linear convolution anymore due to non-linear transformation in norm calculation.
Our overall formulation, however, is not necessarily limited to multiresolution analysis either and we can simply replace downsampling part by corresponding norm computation to support $L_p$ pooling.
This modification however will not retain all the frequency information of the input as it is no longer multiresolution analysis.
We focused on average pooling as it has a clear connection to multiresolution analysis.

\paragraph{Limitations}
We designed wavelet CNNs to put each high frequency part between layers of the CNN.
Since our network has four layers to reduce the size of feature maps, the maximum decomposition level is restricted to five.
This design is likely to be less ideal since we cannot tweak the decomposition level independently from the depth (thereby the number of trainable parameters) of the network.
A different network design might make this separation of hyper-parameters possible.

While wavelet CNNs achieved the best accuracy for training from scratch, its performance with fine-tuning with the ImageNet 2012 dataset is only comparable to FV-CNN, albeit with a significantly smaller number of parameters.
We speculated that it is partially because a more complex model of FV-CNN, but another possibility is that pre-training with the ImageNet 2012 dataset is simply not appropriate for texture classification.
An exact reasoning of failure cases, however, is generally difficult for any neural network models, and our model is not an exception.
We however note that we could not find a publicly available texture dataset at the same scale as the ImageNet 2012 dataset.

We should also point that, while our model improves accuracy a lot when we used textures as only training datasets, the accuracy is still around 60\% for \emph{kth-tips2-b} and 32\% for \emph{DTD}.
This level of accuracy might not be enough yet for certain applications and there is still room for improvement when compared to image classification.

%% file: conclusion.tex

\section{Conclusion}\label{Sec:Conclusion}
%
We presented a novel CNN architecture which incorporates a spectral analysis into CNNs.
We showed how to reformulate convolution and pooling layers in CNNs into a generalized form of filtering and downsampling.
This reformulation shows how conventional CNNs perform a limited version of multiresolution analysis, which then allows us to integrate multiresolution analysis into CNNs as a single model called wavelet CNNs.
We demonstrated that our model achieves better accuracy for texture classification with smaller number of trainable parameters than existing models.
In particular, our model outperformed all the existing models with significantly more trainable parameters when we trained each model from scratch.
A wavelet CNN is a general learning model and applications to other problems are interesting future works.